\documentclass[journal,twocolumn,10pt]{IEEEtran}
\usepackage{multirow}
\usepackage{amsmath}
\usepackage{enumerate}
\usepackage{amsfonts}
\usepackage{setspace}
\usepackage{amsthm}
\usepackage{amscd}
\usepackage{hhline}
\usepackage[dvips]{graphicx} 
\usepackage{bmpsize} 
\usepackage{epsfig}
\usepackage{epstopdf}
\usepackage{amssymb}
\usepackage{multirow}
\usepackage{setspace}
\usepackage{color}
\numberwithin{equation}{section}

\begin{document}


\title{ 
Emerging Biometrics: Deep Inference\\  and Other Computational Intelligence\\
}

\author{   {S.   Yanushkevich,}    {S.   Eastwood},   {K. Lai}, and {V.   Shmerko}   
\thanks{Authors      
 are with Biometric Technology Laboratory, Department of Electrical and Computer
 Engineering,  University of
 Calgary, Canada, Web: http://www.ucalgary.ca/btlab. E-mail:
syanshk\{sceastwo,kelai,vshmerko\}@ucalgary.ca. 
} 
}



\maketitle

\begin{abstract}

This paper aims at identifying emerging computational intelligence  trends for the design and modeling of complex biometric-enabled infrastructure and systems. Biometric-enabled systems are evolving towards  deep learning and deep inference using the principles of adaptive computing, -- the front tides of the modern computational intelligence domain. 
Therefore, we focus on intelligent inference engines widely deployed in biometrics. Computational intelligence applications that cover a wide spectrum of biometric tasks using physiological and behavioral  traits are chosen for illustration. We  highlight the technology gaps that must be addressed in future generations of  biometric systems.   The reported approaches and results primarily address the researchers who work towards developing the next generation of intelligent biometric-enabled systems.
\end{abstract}



\section{Introduction}\label{sec:}

\IEEEPARstart{C}{}ontemporary biometric-enabled systems 
expect  more flexible, intelligent, and reliable solutions. Despite impressive progress in this area, there are a lot of examples of technology gaps related to Computational Intelligence (CI) between the achieved and required performance, in particular:  

\begin{enumerate}
	
		\item []\hspace{-8mm} \textbf{Ambient intelligent systems}, -- an emergency branch of biometrics focusing on the extension
of human abilities in performance improvement and graceful
degradation
	 \cite{[Buchmayr-2011],[IBM-Watson-Health-2017],[Tistarelli-2011]}.  These systems are parts of the smart city and smart home concepts \cite{[Cook-2010],[Doctor-2005],[Petrolo-2017]}.
		\item []\hspace{-8mm} \textbf{Affective computing}, --   an emergency direction of biometrics aiming at the automated analysis of human affective behavior, such as an intelligent tutoring system with feedback to instructors, \emph{e.g.}, e-coaching, e-teaching, and e-training \cite{[Calvo-DMello-2010]}. Multiple biometric modalities are used in these interactions, such as audiovision, touch, smell, movement (\emph{e.g.}, gesture detection), interpretation of human language commands, and other multisensory
functions \cite{[Turk-2014]}. A special case of affective computing is the e-interviewer that aims to detect the likelihood of deception  \cite{[Abouelenien-2017],[Nunamaker-AVATAR-2011],[Twyman-2014a]}.
				\item []\hspace{-8mm} \textbf{Meta-inference engine,} -- CI-related resources for performance improvement.      
	Rodrigues \emph{et al.} \cite{[Rodrigues-2010]} showed that in some attack scenarios, the multi-biometric systems are as vulnerable as single modality systems. The CI related countermeasures are reported in \cite{[Biggio-2017]}. Scheirer \emph{et al.}  \cite{[Scheirer-2012],[Scheirer-Meta-Theory-2011]} explore the meta-recognition of biometric traits. Meta-analysis was used in traveler risk assessment via watchlist by Lai \emph{et al.} \cite{[Lai-Yanush-2017]}.
\end{enumerate}

In these problems, \textbf{proactive computing is a mandatory requirement.}  Given  state $n$ of a biometric-enabled system, calculate the probability  that the system achieves state $n+j$, where $j=1,2,\ldots ,k$ is the depth of prediction.  In such a formulation, potentially required resources can be activated in advance. To make predictive judgments about some scenario is one of the main requirements to the biometric-enabled infrastructure.

The \textbf{core of these systems are CI techniques and adaptive feedback at various hierarchical levels including the machine-human loop.} These applications involve large and complex collections of hidden variables and uncertain parameters. Various perception aspects of biometrics are periodically surveyed aiming at the identification and updating of trends, achievements, and breakthrough directions.  Keeping the same goals, the proposed survey highlights the role and challenges of CI in biometric-enabled infrastructure. 
For this, the following high demand applications of the highest complexity are chosen:  authentication and risk assessment machines,  ambient assistants, affect detectors,   and synthesizers of biometric traits. We argue that these projections of CI-related problems of biometrics cover a wide spectrum of biometric tasks using physiological and behavior traits.

\textbf{The CI-related angle of viewing of biometrics introduced in this paper is different from the one mentioned in the  surveys,} which focus on particular areas,  modalities, and tasks, such as: 
\begin{itemize}
	\item soft biometrics 
		\cite{[Dantcheva-survey-2016],[Fu-2010],[Goffredo-2010],[Nixon-2015],[Tome-2015]};
\item 
	ambient intelligence systems \cite{[Acampora-2013],[Buchmayr-2011],[Diaz-Rodriguez-2014],[Li-2015],[Tistarelli-2011],[Xefteris-2011]}; 
	\item automatic analysis of facial affect \cite{[Corneanu-2016],[Sariyanidi-2015]}; 
	\item adaptive biometric systems \cite{[Calvo-DMello-2010],[Goumopoulos-2011],[Rattani-2015],[Scheirer-2012],[Scheirer-Meta-Theory-2011]}; 
\item deep learning \cite{[LeCun-Hinton-2015],[Srinivas-2016]};
\item free energy principle for deep learning \cite{[Friston-2010]};
	\item multimodality data fusion \cite{[Dinca-2017],[Lahat-2015],[Lumini-2017],[Nigam-2015]};
	\item anomaly detection (biometric-related tasks) \cite{[Chandola-2009]}; 
	\item distance/similarity measures \cite{[Cha-2007],[Jousselme-2012]}; 
	\item spoken dialog systems \cite{[Young-2013]}; and
	\item synthetic biometrics \cite{[Yanushkevich-2007],[Wang-N-2014]}.
	\end{itemize}

Recent revisions of the biometric system design needs has brought a better understanding of the challenges and the short-comings of the applied CI techniques. In particular, \textbf{emerging biometric applications are analyzed in the surveys dedicated to:}  

\begin{itemize}
	\item authentication machines \cite{[Eastwood-IEEE-J-2015],[Labati-2016]},  e-passports/ID technology \cite{[Avoine-2016]}, 
	\item multimodal interactions \cite{[Turk-2014]}, 
	\item biometric recognition in surveillance scenarios \cite{[Neves-2016]},
	\item health, ambient intelligence, and security \cite{[Faundez-Zanuy-2013]}, 
	\item forensics  \cite{[Jain-2015],[Lai-Yanush-2017]}, 
	\item affect-aware computer applications \cite{[Calvo-DMello-2010],[Corneanu-2016],[Zeng-2009]}, and 
	\item smart cities \cite{[Petrolo-2017]}.
\end{itemize}

These and other surveys have very different foci. In this paper, we target the CI-related advances in the biometric technologies that interlace with and benefit from the latest achievements in cognitive science. The CI is the major foundation upon which the next generation of biometric-enabled machines is being stemmed. This generation includes e-teachers that increase learning efficiency,  e-coaches that address medical applications, e-interviewers targeting security applications, etc. 
The CI related view is timely and in great demand, as biometric technologies tasks became more complicated and are often formulated in terms of knowledge inference and machine learning. 

The variability of biometric data, unstable nature of the measured biometric phenomena, uncertainty in decision-making,  strong requirements for the reliability of biometric-enabled systems, interdisciplinary of technologies, and deep social embedding, -- all of these represent a fraction of the multiple factors that prompt the seeking of new CI processing approaches and paradigms.  The example chosen in this paper demonstrates an inference engine for an application scenario that utilizes biometric traits.

This paper is organized as follows. Section \ref{sec:What-kind-computation-we-need} poses the question of what kind CI techniques are needed in biometrics. It is followed by 
$i)$ Preliminary exploration and assessment (\ref{subsec:Preliminary}),
$ii)$ Taxonomical sketch of
inference engines (\ref{subsec:Sketch}), 
$iii)$ Inference engine (\ref{subsec:Inference}), and
$iv)$ Technology gap navigator (\ref{subsec:Technology-gap}).
Beyond the overview of the CI approaches to  biometrics, this paper provides perspectives and guidelines on how to apply the latest advanced CI techniques such as deep inference for frontier biometric applications (Section \ref{sec:Frontiers-appl}). Four examples are provided: 
1) Authentication and risk assessment machines (\ref{subsec:Authentication}),
2) Ambient assistants (\ref{subsec:Ambient}),
3) Affect analyzers (\ref{subsec:Affect}), and
4) Synthesizers of biometric traits (\ref{subsec:Synthesizers}).
		Section \ref{sec:Summary-conclusion} concludes this survey.
			
 The following standard abbreviations of inference engine components are used throughout this paper:

\begin{tabular}{lll}
\hline
\textbf{BN}   &$-$& Bayesian Network;\\
\textbf{CI}   &$-$& Computational Intelligence;\\
 \textbf{CNN} &$-$& Convolutional Neural Network;\\
 \textbf{CRF} &$-$& Conditional  Random Field;\\
 \textbf{DBM} &$-$& Deep Boltzmann Machine;\\
 \textbf{DBN} &$-$& Dynamic Bayesian Networks;\\
 \textbf{DSN} &$-$& Dempster-Shaffer Network;\\
 \textbf{HMM} &$-$& Hidden Markov Model;\\
 \textbf{MRF} &$-$& Markov Random Field;\\
 \textbf{PCA} &$-$& Principle Component Analysis;\\
 \textbf{RBM} &$-$& Restricted Boltzmann Machine;\\
 \textbf{SVM} &$-$& Support Vector Machine.\\
\hline
\end{tabular}

\section{What kind of inference and CI is needed?}\label{sec:What-kind-computation-we-need}

In a very general form, the \textbf{inference problem is formulated as follows:
given a set of noisy or ambiguous biometric measurements, infer the likely state of this biometric trait. }
It is impossible to make these inferences  with complete certainty, but one can at least try to obtain the most probable state of a hidden part of the biometric trait, within a chosen model and measurements. Note that inferring the most probable state is an \textbf{optimization problem} and can be represented in various forms depending on the model, such as minimum energy, minimum potential, and minimum distance.

Inference addresses computing the posterior probability distribution of certain variables given some value-observed variables as evidence.
It is a well-accepted paradigm in certain biometrics that features of interest in biometric traits should be identified by deep inference and learning technologies. 

In this section, we  provide  a more concrete sense to what we
mean when we speak of ``inference'' and ``CI'' in biometrics.
From the above perspectives, we need to revise the CI-related methodologies and provide a taxonomical view of the available resources.

\subsection{Preliminary exploration and assessment}\label{subsec:Preliminary}

The view of contemporary biometrics in the light of CI technology requires some preliminary exploration and assessment of this field.  
This assessment infers that the \textbf{contemporary biometric device or system is part of a more complex organism whose resources are distributed over both the physical and digital/virtual world.} 

The biometric-enabled systems become embedded in social infrastructure. Examples include intelligent surveillance networks \cite{[Zouaouia-2015]}, the tracking and recognition of persons of interest \cite{[Feris-2014]}, and performing authentication and risk assessment of individuals in transportation hubs \cite{[Lai-Yanush-2017],[Lai-Eastwood-2018]}. 
Biometrics are an integrated part of the smart home \cite{[Cook-2010],[Doctor-2005]} and smart city \cite{[Petrolo-2017]} concepts. In response to these trends, there is a demand for the most advanced CI techniques and technologies. In addition, the privacy issues of biometrics contributed to design challenges. 
  However, privacy in the age of biometrics is out of this paper's scope.

  \subsection{Sketch of the taxonomical revision of inference engines} \label{subsec:Sketch}

 In this sub-section, the taxonomical projections on inference engines are revised. The core of the inference engine is a graphical model that provides a framework for dealing with high-dimensional probability distributions. In such a model, the nodes of the graph represent the variables on which the distribution is defined, the edges between the nodes reflect their probabilistic dependencies, and a set of functions relating nodes and their
neighbors in the graph are used to define a joint distribution over all of the variables based on those dependencies. 
 Typical inference problems include finding the marginal probability of a task-relevant variable, or finding the most probable explanation of the observed data.

Probabilistic task formulation in contemporary biometric-enabled systems  reflects the real world more realistically. For example, in e-borders, the question \texttt{``Is this traveler on the watchlist?''} \cite{[kn:Bigo-CEPS-2012]} is reformulated as \texttt{``What is the risk (or cost) of the traveler being wrongly matched/non-matched to the watchlist?''} \cite{[Eastwood-2014-b],[EC-SmartBorders-2014]}. This means that the watchlist check by a human operator  is replaced by a \emph{watchlist-based inference} of risks, which is provided to a human operator.

Conceptually, the inference engine consists of two types of inference models \cite{[DudaBook2012],[Frey-2005],[Koller2009]}:
1) \emph{Discriminative models} that \textbf{predict the distribution of the output given the input}, such as linear regression 
(the output is a linear function of the input  plus Gaussian
noise), and Support Vector Machines (SVMs) (the binary class variable is Bernoulli distributed with a probability given by the distance from the
input to the support vectors), and
2)  \emph{Generative models} that account for
all of the data and provide a more general way to
combine the preprocessing task and the discriminative task. 

In this paper, the focus of our interest are the models that exploit the statistical properties of biometric traits more efficiently by \emph{learning}, especially ones that capture high-order dependencies, --   \emph{deep inference} models. Such models are useful not only for discriminative tasks but also for providing adaptive priors for the generative tasks. They are able 
to $i)$ adapt to the input data, and $ii)$ decompose the problem
of learning hierarchical nonlinear systems into a sequence of simpler learning tasks.

 Our survey is based on the criterion that the \textbf{goal is not to choose the model that is the best fit, but to choose a model that fits the data well and is consistent with prior knowledge.}
While a rigorous mathematical and taxonomical vision of the inference and learning algorithms is introduced by Frey and Jojic \cite{[Frey-2005]},  here we summarize a general landscape of inference engines as follows:

\begin{enumerate}
\item []\hspace{-8mm} \textbf{$-$ Energy-based inference}  which associates a scalar energy with each configuration of the
variables of interest.  The idea is to approximate the true posterior distribution
by a simpler distribution, which is then used for
making decisions.  We recommend the survey by Frison \cite{[Friston-2010]} for details.
\item []\hspace{-8mm} \textbf{$-$ Deep learning inference.} The idea is
	to exploit the property that  natural signals such as  biometric traits, are  compositional hierarchies. That is, higher-level features can be obtained by composing lower-level ones. The fundamentals of deep learning inference are developed in \cite{[Hinton-Sejnowski-1986],[Hinton-2006],[Hinton-2006-a],[Neal-Hinton-1998],[LeCun-Hinton-2015]}. 
\item []\hspace{-8mm} \textbf{$-$ Biologically inspired inference} such as evolutionary algorithms for generating synthetic fingerprint images \cite{[Cho-2007]},   skin formation modeling \cite{[Kuecken-2004]}, and features of plastic surgery \cite{[kn:Bhatt-Plastic-Surgery-2013]}.
	\item []\hspace{-8mm} \textbf{$-$ Multimetric inference}   based on the concept of probability propagation in causal networks. The main idea is to use a uniform graphical platform that combines various metrics such as point probability estimation in naive Bayesian Networks (BN) \cite{[Koller2009]}, interval probability BNs \cite{[Lai-Eastwood-2018]}, Fuzzy Probability BNs \cite{[Baldwin-2003]},   Dempster-Shafer BNs \cite{[Delmotte2004],[EastwoodCIDS2015],[Jirousek2012],[Simona-2008]}, and their Dezert-Smarandache extensions \cite{[Smarandache-Dezert-Tacnet]}.
\item []\hspace{-8mm} \textbf{$-$ Hybrid approaches}  such as message-passing inference that combines inference possibilities of both the BNs and MRFs, with example applications to soft biometrics \cite{[Chhaya-2012]}, as well as  Dynamic Bayesian Networks (DBNs) \cite{[Koller2009]}.
\item []\hspace{-8mm} \textbf{$-$ Meta inference} such as meta-recognition for performance prediction  \cite{[Scheirer-2012],[Scheirer-Meta-Theory-2011]} and selecting a classifier from an ensemble for recognition improvement  \cite{[Cruz-2015]}. Traditional classification algorithms are very successful in meta-learning,
in particular,  SVMs  \cite{[Miranda-2014]} and RBMs \cite{[Papa-2015]}.
\end{enumerate}

The above inference engines exercise the learning paradigm in different manners.
For example, Bayesian inference stipulates how  learners should update their beliefs in the light of evidence. 
A Markov network, also called MRF, is similar to a BN in its representation of dependencies. The difference between them is that Bayesian networks are directed acyclic graphs, whereas MRFs are undirected and may have cycles. Thus, an MRF can represent certain dependencies that a BN cannot (such as cyclic dependencies). On the other hand,  a BN can describe  induced dependencies. Learning in energy-based models such as MRFs corresponds to modifying that energy function to have desirable properties \cite{[Koller2009]}.

The \emph{power of the inference engine} refers to its ability to capture high-order dependencies in the data. For example, deep learning methods construct hierarchies composed of multiple layers by greedily training each layer
separately using unsupervised algorithms \cite{[Hinton-2006],[Hinton-2006-a]}.   These hierarchical representations encode information in a
low-level to high-level manner.  Deep learning is a paradigm that learns multi-layered hierarchical representations from data.

Good performance is reported in various tasks, in particular, in silicon face mask detection \cite{[Manjani-2017]} and hair style recognition \cite{[Proenca-2017]}. 
This is achieved due to the ability of algorithms to $i)$ adapt to the input data, $ii)$ build recursively  hierarchies using
unsupervised algorithms, and $iii)$ represent the problem of learning as a sequence of simpler learning tasks.

\subsection{Inference engine}\label{subsec:Inference}

More detailed visions of the inference engine    requires a notion of the \emph{inference engine navigator}, which is  a set of indicators that characterized the CI mechanism. Ideally, an indicator should be reliable, sensitive, quantifiable, and should provide information on the nature of the inference methodology, sources of  risks, hazards,  requirement to the input data, their preprocessing, encoding, as well as the output. 
The key indicators that provide a way for choosing an appropriate inference algorithm, and for  predicting and interpreting  the results, are listed in  Table \ref{tab:Indicator}
\cite{[Frey-2005],[Koller2009]}.

\begin{table*}[!hbt]
\caption{Inference engine navigator as a representative set of key indicators.}
\label{tab:Indicator}
\begin{small}
\begin{center}
\begin{tabular}{|lc|l|}
\hline
  &\multicolumn{1}{c|}{\textbf{Indicator}}&
  \multicolumn{1}{c|}{\textbf{Content of the inference and learning indicator}}
  \\\hline
I1. &\begin{parbox}[h]{0.15\linewidth}{
\vspace{1mm} Graph type
 \vspace{1mm}}
\end{parbox}&
\begin{parbox}[h]{0.75\linewidth}{
\vspace{1mm} Graphical representations of the scenario that is being modeled fall into two main categories: causal or non-causal. Causal models are denoted using directed graphs that do not contain any directed loops: a ``directed acyclic graph". Non-causal models are denoted using the more general non-directed graphs. 
\vspace{1mm}}
\end{parbox}\\   \hline
I2. &\begin{parbox}[h]{0.15\linewidth}{
\vspace{1mm} Cost function
 \vspace{1mm}}
\end{parbox}&
\begin{parbox}[h]{0.75\linewidth}{
\vspace{1mm} The ``cost function" associated with a model returns a number when presented with a specific scenario, evidence, or choices. 
\vspace{1mm}}
\end{parbox}\\   \hline
I3. &\begin{parbox}[h]{0.15\linewidth}{
\vspace{1mm}  Approximate p.d.f.
 \vspace{1mm}}
\end{parbox}&
\begin{parbox}[h]{0.75\linewidth}{
\vspace{1mm} A p.d.f. can be in the form of probability interval distributions (closed intervals in place of point probabilities) and fuzzy probability distributions (triangular fuzzy numbers in place of point probabilities).
\vspace{1mm}}
\end{parbox}\\   \hline
I4. &\begin{parbox}[h]{0.15\linewidth}{
\vspace{1mm} Conflict resolving
 \vspace{1mm}}
\end{parbox}&
\begin{parbox}[h]{0.75\linewidth}{
\vspace{1mm} Models and metrics that are ``conflict resolving'' have the ability to handle apparent contradictions and disagreements in the initial and observed data. 
\vspace{1mm}}
\end{parbox}\\   \hline
I5. &\begin{parbox}[h]{0.15\linewidth}{
\vspace{1mm} Metrics
 \vspace{1mm}}
\end{parbox}&
\begin{parbox}[h]{0.75\linewidth}{
\vspace{1mm} The ``metric'' refers to certain quantities used to quantify uncertainty. Metrics include point probabilities, interval probabilities, fuzzy probabilities, DS models, and DSm models. Metrics can be used in parallel to generate a more comprehensive depiction of the scenario. 
\vspace{1mm}}
\end{parbox}\\   \hline
I6. &\begin{parbox}[h]{0.15\linewidth}{
\vspace{1mm} Learning/Training
 \vspace{1mm}}
\end{parbox}&
\begin{parbox}[h]{0.75\linewidth}{
\vspace{1mm} ``Learning'' refers to a process used to generate the parameters for a model, which are chosen using available statistical data. 
\vspace{1mm}}
\end{parbox}\\   \hline
I7. &\begin{parbox}[h]{0.15\linewidth}{
\vspace{1mm} Dynamic
 \vspace{1mm}}
\end{parbox}&
\begin{parbox}[h]{0.75\linewidth}{
\vspace{1mm} Dynamic models track the evolution of a system with respect to time. Probability based dynamic models include DBNs, CFNs, 
and Markov chains.
\vspace{1mm}}
\end{parbox}\\   \hline
\end{tabular}
\end{center}
\end{small}
\end{table*}

\paragraph{\bf Graph type}
 Visualization of a given  modeling scenario is a well determined stage \cite{[Koller2009]} that is crucial in CI-related processing. Graphical models provide a natural way to represent dependencies of variables. They
specify a complete joint probability distribution function (p.d.f.) over all the variables; this type of model, such as Bayesian networks,  use causal (directed) graphs, while MRF and CRFs  use non-causal (undirected) graphs that represent symmetric dependencies.

Probabilistic inference in graphical models pertains to computing a conditional probability distribution over the values of  the hidden or unobserved nodes, given the values of evidence or observed nodes.

Given the joint p.d.f., all possible inference queries can be found by marginalization (summing out irrelevant variables).
The factor graph has  emerged as a unified model of directed   and undirected graphs  \cite{[Kschischang-2001]}. For example, recently Laar and Vries \cite{[Laar-2016]} have reported the application of a message-passing algorithm in ambient computing such as hearing loss compensation.

An alternative approach addresses \emph{graph learning} or \emph{network topology inference} \cite{[Holbert-2015]}. This approach is briefly explained as follows.
In most modeling scenarios, the graph structure is known. However, in many tasks such as facial expression or gait recognition, it is reasonable to suppose that graph topology is not fixed, it is varying with time. This variability can be described in terms of transition probabilities, probability distributions, and is modeled using, for example, semi-Markov processes.  In this case, the graph topology must be inferred by the topology inference algorithms. Shuman \emph{et al.} \cite{[Shuman-2013]} analyzed  challenges of signal processing on graphs.  Meena \emph{et al.} \cite{[Meena-2017]} reported results on facial expression recognition.
Properties of varying graphs (hidden structure and dynamic relationships) were studied in \cite{[Kalofoliast-2017]}; the paper also proposed an optimization framework using factor-graphs and a message-passing algorithm.

\paragraph{\bf Cost function}
  Many approximate inference techniques can be viewed as a process of minimizing a cost function, which measures the accuracy of an approximate probability distribution \cite{[DudaBook2012],[Frey-2005],[Koller2009],[RussellBook2009]}. 
The free energy principle \cite{[Friston-2010]} has been applied to recognition and person re-identification. For example, free energy  score computing was applied to discriminative classifiers \cite{[Perina-2012]}.  If the MRF model is chosen, there are two strategies for optimization: graph-cuts and  message-passing based on factor graphs.  In \cite{[Komodakis-2011]}, the MRF energy is minimized by the so-called tree-reweighted message-passing algorithm. The essence of this approach is to decompose the original MRF optimization problem, which is NP-hard, into a set of easier MRF subproblems, each one defined on a tree. 
High-order MRFs called the \emph{gated} MRFs, have been  developed in \cite{[Ranzato-Hinton-2013]}: one of  two sets of latent variables were used in order to create an image-specific energy function that models the covariance structure.  The authors argued that such MRFs  can be used as the front-end of a standard deep architecture, often called a deep belief network  \cite{[Hinton-2006]}.

Note that the topology inference problem is formulated in terms of energy functions such as the Dirichlet energy, Flobenius energy, and the cumulative energy  \cite{[Kalofoliast-2017]}.

\paragraph{\bf Approximate p.d.f.} 

 In approximation techniques, the distribution of parameter values before any data is examined is called the \emph{prior} distribution; the \emph{posterior} distribution is the conditional distribution of the parameter given the data.

Inference in most models, including  Boltzmann machines, is NP-hard in general. As a result, approximations are often necessary. Even in cases, in which the complexity of the exact algorithms is manageable, it can be reasonable to consider approximation procedures. The criterion of model approximation is useful in choosing the inference engine. For example, the MRF approximation  model proposed in \cite{[Komodakis-2011]} is constructed based on a set (a ``bundle'') of subgradients of the objective function, which is continuously refined throughout the minimization algorithm.
Sutskever and Hinton \cite{[Sutskever-Hinton-2008]} showed that deep belief networks can approximate any distribution over binary vectors to an arbitrary level of accuracy, even when the width of each layer is limited to the dimensionality of the data. 

As a rule, the Gibbs distribution approximates the real distribution (uncertainty) if the energy cost function is chosen. Scheirer \emph{et al.} \cite{[Scheirer-2012],[Scheirer-Meta-Theory-2011]} proposed a new statistical classifier based upon the Weibull distribution that produces accurate predictions of image recognition success or failure on a per instance  basis. This approach is called meta-recognition, as it allows for adjustments of the recognition decisions using post-recognition similarity score analysis. In meta-recognition, the so-called tail approximation borrowed from extreme value theory is used. In addition, it provides a robust normalization for the system's non-match scores.

Bayesian inference  faces a  class of settings,
where the likelihood function is not completely known and where exact
simulation from the corresponding posterior distribution
is impractical or even impossible. 
Such settings call for practical
if cruder approximations methods. Approximate Bayesian computation, or \textbf{likelihood-free inference
algorithms}, were developed  for performing Bayesian inference  (approximations to posterior distributions) without the need for
explicit evaluation  of the model likelihood function \cite{[Marin-2012],[Turner-2012]}.
These likelihood-free techniques can be used in scenarios where the model of interest has become intractable or an ill-behaved likelihood function.

\paragraph{\bf Conflict resolving} 
Facts  about the same real-world object are collected from various sources, \emph{e.g.}, facial expressions from the visual band and infrared cameras, and gait traits from various cameras.  The objective of a conflict resolution algorithm, also known as truth discovery, is to distinguish between true and false patterns. This could  be achieved by assigning a correctness score to each pattern, such that the highest scores are assigned to the true patterns.

Given the prior distributions and the conditional
probabilities,  Bayesian inference offers a complete, scalable, and theoretically justifiable approach for various tasks, such as prediction and data fusion \cite{[DudaBook2012],[Koller2009]}.  However, in real scenarios, such complete knowledge is difficult or impossible  to obtain. BNs become numerically unstable when presented with contradictory information. 

The Dempster-Shafer (DS) evidence theory
 allows for the representation of both uncertainty and imprecision, and
can effectively deal with missing information and complimentary hypotheses.
In DS theory \cite{[Shafer1987]} and the related transferable belief model \cite{[Delmotte2004]}, the uncertainty and imprecision are represented via the notion of confidence values that are committed to a single or a union of hypotheses. These possibilities are  useful in practice \cite{[Mezai-Hachouf-2015],[Nguyen-2015]}, including the fusion of local and global matching scores \cite{[Kisku-2009]}.
 However, DS theory, as well as Bayesian inference, fail to model conflicts in data that may arise, for example, between information sources. Specifically, Bayesian inference assumes that all sources provide bodies of evidences using the same objective and universal interpretation of the phenomena under consideration; therefore, it cannot handle conflicts.  
In DS inference, the conflict resolving mechanisms are weak and often lead to false conclusions especially for a large number of conflicts.  

The Dezert-Smarandache  (DSm) theory of reasoning with plausible and paradoxical cues helps to overcome these limitations \cite{[Smarandache-Dezert-Tacnet]}. The DSm theory demonstrates good results in the practice of multi-source fusion \cite{[Zouaouia-2015]}. One of the reasons is that DSm can distinguish sources involved in the fusion process,  with respect to reliability (an objective property) and importance (a subjective property).

\paragraph{\bf Metrics} 
Useful recommendations on choosing an 
appropriate measurement metric, including DS evidence theory, are given in \cite{[Jousselme-2012]}. 
Samples of the most common  measuring approaches are as follows:
\begin{itemize}
	\item Point probabilities utilized in classical models such as BN, MRF, DBM, and PCA.
		\item  Probability interval BNs \cite{[DeCampos1994]}.
			\item Credal networks \cite{[Karlsson2011]}.
\item Fuzzy BNs \cite{[Baldwin-2003]}.
\item DS BNs \cite{[EastwoodCIDS2015]}.
\end{itemize}
It is reasonable to combine these and other metrics at the unified graphical platform such as causal networks \cite{[EastwoodCIDS2015],[Lai-Eastwood-2018]}.  The strength of this approach is that it introduces the uncertainty in different metrics. 
For example, the integration of the fuzzy metric in regression analysis \cite{[Wang-2000]},  SVM \cite{[Hao-2008]}, Bayesian networks \cite{[Baldwin-2003]}, and RBMs \cite{[Chen-2015]} provides a competitive performance in data representation capability and robustness in order to cope with various types of uncertainties.

Biometric trait processing using information theory measurements is of special interest.
There are a lot of efforts on the application of information-theoretical measures for the performance evaluation of a biometric system, such as the information content of biometric traits, for example iris \cite{[Daugman-2016]}, and the information gained through biometric systems \cite{[Takahashi-2014]}. 
Zhang \emph{et al.} \cite{[W_Zhang-2011]}  proposed the use of an information-theory approach for face photo-sketch recognition. 
In \cite{[Lim-2016]}, an entropy-measuring model for biometric verification systems has been proposed.  Specifically, authors  showed that the imposter distribution  leaks information about the randomness of the biometric representations. This work is closely related to meta-learning \cite{[Miranda-2014],[Papa-2015]} and meta-recognition \cite{[Scheirer-2012],[Scheirer-Meta-Theory-2011]}.

A study by Wang \emph{et al.} \cite{[Y_Wang-2012]} offers a generalized framework for the analysis of biometric systems, including privacy leakage, in terms of information theory.  In particular, they specified a fundamental trade-off between the user's privacy (provided personal biometric information) and the user's security (probability that the adversary can falsely authenticate
as a genuine user) for a multiple biometric system.
Entropy can also be useful in e-interviewer development,  as shown in \cite{[Wu-2015]}.

Other applications of information theory in biometrics include,  for example,  measuring the  similarity between two
probability distributions (e.g. true and approximate). The relative entropy (also known as the Kullback-Leibler divergence) has been used by \cite{[Frey-2005],[Koller2009]}.

\paragraph{\bf Learning/Training}
  Learning can be thought of as inferring plausible models to explain observed data.  
	It is well documented that 	models which include a latent (hidden-state) structure may be more expressive than fully observable models \cite{[Hinton-Sejnowski-1986],[Hinton-2006],[Hinton-2006-a]}. The ultimate goal of unsupervised learning is to discover representations that
parametrize a lower dimensional yet highly nonlinear manifold, and hence capture the intrinsic structure of the input data. This structure is
represented through features also called latent variables in  models. For example, HMMs and DBMs use hidden states to model observations; their tasks  are formulated in terms of  generative probabilistic functions. A limitation of generative models is that observations are assumed to be independent given the values of the latent variables.

Another discriminative model well suited for   language processing, gesture  and emotion recognition, in particular, is the Conditional Random Field (CRF). They make use of arbitrary feature-vector representations of the observed data points. Various modifications of CRFs are known, such as the dynamic CRF used to learn the hidden dynamics between
input features \cite{[Baltrusaitis-2013]} and the continuous CRF used to model the affect continuously  \cite{[Ramirez-2011]}.

Hybrid models such as  \emph{Bayesian deep learning}, were recently proposed by Wang and Yeung \cite{[H-Wang-2016]}.

Depending on the model type, different instabilities in learning are observed. For example,  when training a DBM, approximation instability addresses the effect of noisy gradients. 
Learning with both BNs and DBNs involves two aspects:  learning the structure (graph topology) and  learning the parameters  (conditional probability tables) for each variable. It is possible to model scenarios with full and partial observability (e.g., missing  data).

Meta-learning is aimed at building a self-adaptive learner with the ability to improve its biases dynamically, thus increasing its efficiency through experience. For this, meta-knowledge (knowledge about knowledge) must be accumulated and analyzed. The key question in this process is what kind of meta-features are suitable in a given scenario (dataset, inference mechanism, algorithm, task). Traditional classification algorithms such as  SVMs  \cite{[Miranda-2014]} and RBMs \cite{[Papa-2015]} were shown to be successful at meta-learning. 


\paragraph{\bf Dynamic networks} 
 
Biometric tasks often require the processing of an audio data stream or a surveillance video stream using a feedback loop. Such problems can be formulated as control of a non-linear dynamic process. This is a typical iterative affective process aiming at the prediction of the identified object's dynamics (human walking, talking, running, or facial expression). In such a formulation, the two phases should be distinguished: recognition of an object of interest, and control of the process. 

Another critical factor in some scenarios such as gait analysis \cite{[Xiao-2013]}, face identification in video \cite{[NIST_Face-In-Video-2017],[Zhu-2016]}, facial expressions recognition \cite{[Nixon-2015]}, and pulse estimation \cite{[Hsu-2017]}, is the time factor. Dynamical properties of the Restricted Boltzman Machine (RBM) for processing video sequences have been demonstrated by  Sutskever and Hinton  \cite{[Sutskever-Hinton-2008]}.  The DBMs have been used in \cite{[Souza-2017]} for robust fingerprint spoofing attack detection. The joint DBM for audio-visual person identification have been proposed in \cite{[Alam-2017]}.  

DBNs are an extension of BNs to model dynamic processes \cite{[Koller2009]}. A DBN consists of a series of time slices that represent the state of all the variables at a certain time. Usually, DBNs
are restricted to have directed links between consecutive temporal slices, known as a first-order Markov model. They can be seen as a generalization of Markov chains and hidden Markov models (HMMs).

\subsection{Technology gap navigator}\label{subsec:Technology-gap}

Lai \emph{et al.} \cite{[Lai-Eastwood-2018]} proposed to use the technology gap methodology developed by the U.S. Pacific Northwest National Laboratory  \cite{[Hartman-Technology-Gap-Analysis-2005]}, for the purpose of biometric-enabled infrastructure. The following key drivers are defined: 
the \emph{opportunistic driver} (whether or not suitable signatures can be developed), the \emph{mature driver} (whether or not a suitable deployment scenario can be developed), and the \emph{development driver} (whether or not a suitable measurement method can be developed). The two sides of the gap should be specified: the technology state-of-the-art and the required technology for a given problem. Some examples of those are provided below:	
	
1) In the technology gap navigator for large biometric-enabled infrastructure, known as ``The Checkpoint of the Future''  \cite{IATA-Checkpoint-future},  the basic drivers include risk assessment,   technology,   operations, and their components or modules. The components include passenger data, known traveler data (pre-screening),  identity management, behavior analysis, alternative measures for unpredictability and deterrence, etc. Examples of technology gap navigators are also provided in \cite{[kn:Bigo-CEPS-2012],[EC-SmartBorders-2014],[Trochu-2013]}.

 2) In state-of-the-art  facial identification of non-cooperative individuals (such as surveillance in mass-transit systems) \cite{[NIST_Face-In-Video-2017],[Goswami-2017]}, one aspect of the technology gap states that contemporary surveillance tools  provide a large spectrum of possibilities for law enforcement in mass-transit hubs. 
Another aspect is highlighted in \cite{[Eastwood-IEEE-J-2015]}:  the authentication and risk assessment mechanisms should be significantly improved.  This technology gap states that low quality facial traits (e.g. from surveillance cameras) cannot be used for the purpose of identifying a person of interest via automated watchlist screening because of the poor performance of facial recognition tools.

3) The gap in  deep inference technology using the RBM has been specified, in particular, by Chen \emph{et al.} \cite{[Chen-2015]}. It involves   
$i)$ the model selection (such as how many units are in a hidden layer, and how many hidden layers),
$ii)$ the setting problems (such as Gibbs step, learning rate, and batch learning), and
$iii)$ the computational cost of training (needs to be speed up using more efficient optimization techniques including various metrics).

\section{Frontiers in applications}\label{sec:Frontiers-appl}

To demonstrate the CI related challenges in biometric-enabled systems and infrastructures, we chose the following
 practical applications:
 \begin{itemize}
\item [--] \textbf{Authentication and risk assessment machines} for mass-transit hubs and large public events. These machines  operate under the umbrella of specific supporting infrastructure \cite{[Labati-2016],[Lai-Yanush-2017],[Lai-Kanich-2017]}.
\item [--] \textbf{Ambient assistants} aims at social issues of critical importance: care management  based on new cognitive paradigms \cite{[IBM-Watson-Health-2017]}. 
\item [--] \textbf{Affect detectors} for affect-aware interfaces that aim at automated detection and intelligently respond to users' affective states in order to increase usability and effectiveness \cite{[DMello-2015]},  and 
\item [--] \textbf{Synthesizers of biometric traits,} useful countermeasure technologies \cite{[Cappelli-2003],[Galbally-2014],[Yanushkevich-2007]}; they may prove to be useful for generating data for deep learning.
	\end{itemize}

The common features of these applications are as follows: $i)$ they are  CI related; $ii)$ they  reflect challenges encountered in biometric research, and $iii)$ synthetic biometrics are in great demand in CI related techniques  for learning, testing, and attack countermeasures.

\subsection{Authentication and risk assessment machines}\label{subsec:Authentication}

\paragraph{\bf Tasks}  An authentication and risk assessment machine aims at the identification or verification of a subject, as well as his/her risk assessment, using available sources, \emph{e.g.}, watchlist check.
The framework of this task \cite{[Lai-Kanich-2017]} involves 
three types of human identity:  \emph{attributed} (name, date, and place of birth); \emph{biometric}  (such as face, iris, fingerprint, retina,   gait,  dynamic signature, and DNA profile); and \emph{biographical}  (life events  including details of education,  employment,   marriage, mortgage, and property ownership).

\paragraph{\bf State-of-the-art} It is well documented \cite{IATA-Checkpoint-future,[EC-SmartBorders-2014],[NIST_Face-In-Video-2017],[Robertson-2017],[Trochu-2013]} that the key trend of e-borders is the integration of intelligent support at all levels of surveillance, control, and decision-making. Evidence accumulation and  risk assessment machines are the critical components of this trend \cite{[Trochu-2013],[kn:Frontex-2012-Best-Pracetic]}. They are mandatory in border crossing  checkpoints, airports, and seaports, and will be included in the future  transportation systems and mass transit hubs \cite{[kn:Fiondella-security-mass-transit-systems-2012]}.

In the area of risk assessment, significant progress has recently been reported in tasks such as:
\begin{itemize}
\item [$-$]  watchlist check using surveillance face images  \cite{[Zhu-2016]}.
\item [$-$]  screening technology \cite{[Sgroi-2015]}.
\item [$-$]   proactive  post-recognition score analysis \cite{[Scheirer-2012]}. 
\item [$-$]  face verification from surveillance video frames \cite{[Goswami-2017]} achieved by using DBM, entropy-based selection, and the fusion of abstract and low-level features.
\item [$-$]  iris recognition based on HMM model and information theory measures  \cite{[Daugman-2016]}.
				
\item [$-$] fingerprints, finger-vein, and finger knuckle
patterns that were simultaneously acquired and employed for
more reliable biometrics identification \cite{[Kumar-2016],[Labati-2013]}.
\end{itemize}

  In modeling or thwarting the attacks on biometric systems, CNNs were recently applied for fingerprint liveness detection \cite{nogueira-2016}, as well as for iris, face, and fingerprint spoofing detection \cite{menotti-2015,[Souza-2017]}. Face scrambling as privacy protection during Internet-of-Things-targeted image/video distribution was suggested in
\cite{[Jiang-2016]}. A new type of identity attack in automated border control infrastructure, the so-called  double-identity fingerprint and double-identity face image,  was studied in \cite{[Ferrara-2017]}.  
This type of attack addresses high-risk border crossing scenarios. 
	The authors of \cite{[Manjani-2017]} have  introduced  a novel detector of the facial silicone mask presentation attack based on a multilevel deep dictionary via greedy learning. In \cite{[Proenca-2017]},  a multi-layered (hierarchical) MRF for facial hair style analysis was utilized. This MRF model does not use high
order cliques but still  reaches globally coherent
solutions.  In \cite{[Bustard-2014]}, the mitigation of targeted biometric impersonation has been proposed. Targeted impersonation is defined as a method of spoofing the biometric traits.

Soft biometrics are becoming a priority in some applications such as terrorist countermeasures, security screening, and tracking persons of interest via surveillance networks. 	
Soft attributes have a semantic interpretation, such as ``tall'', ``young'',  ``female''. In the survey  \cite{[Dantcheva-survey-2016]},  an inference engine is for the prediction of soft biometric traits. It employs relationships between biometric modalities and physical attributes which can be inferred from observed data. For example, the body mass index can be predicted from face images \cite{[Wen-2013]},  age can be derived from gait, and skin disease from facial images.

Public safety and security based on surveillance networks has an urgent need of CI techniques for human identification using semantic descriptions (\emph{e.g.}  ``pointy nose'', or ``puffy lips'') in eyewitnesses verbal statements  \cite{[Almudhahka-2018]}. In  \cite{[Karpathy-2017]},  CI techniques are used for the more general task of automated regeneration of image descriptions.

\paragraph{\bf Technology gap} 
 Regarding  technology gaps in authentication, 
 it has be suggested that three types of identity proofs (attributed such as by a document, acquired, \emph{e.g.,} a password or something a subject knows or reveals during an interview, and the biometric one),  must be combined in automated screening. In e-borders, the e-passport introduces two types of identity: an attributed and a biometric one \cite{[IATA-ABC-Guide-2015]}. 
		Emerging solutions are needed for risk assessment to mitigate the  impacts of various unwanted effects of biometric traits such as impersonation   \cite{[Lai-Eastwood-2018]}.
A useful representation of the technology gap in the area of deception detection is given in \cite{[Weinberger-2010]}  (despite it being eight years old, it is mostly still true today).

In on-line criminal investigations, short-term watchlists can be deployed in transportation hubs for searching for suspects. It is based on detailed descriptions provided by eyewitnesses or compiled from images captured by surveillance cameras. This problem is known as the detection of \emph{semantic visual attributes}. 
Such watchlists contain the  fine-grained attributes of persons of interest.  Attribute detectors aim at the extraction of 1) facial features such as bald, beard, mustache,  color of hair,  hat, sunglasses, eyeglasses,   skin tone,   and gender, and 2) torso attributes such as  clothing color, patterned, and solid. An experiment using a large set of surveillance cameras monitoring metro turnstiles was conducted in order to evaluate face capture and attribute-based person identification \cite{[Feris-2014],[Wang-2016]}. Facial sketches were investigated as helpful data to be added to the short-term watchlists \cite{[Zhang-S-2017],[Zhang-L-2015]}. 
Authentication and risk assessment are also mandatory mechanisms in the smart home \cite{[Cook-2010],[Doctor-2005]} and smart city \cite{[Petrolo-2017]}.

Designing biometric-enabled security for smart cities is another challenging problem.  Surveillance networks with visible band  cameras  are useful only in daytime. In a no-light environment, \emph{active}  IR band  cameras should be used. Near IR illumination provides    acceptable conditions for the restoration of near IR images.   For example, 98\%, 94\%, and 76\%
accuracies for 60m, 100m, and 150m screening near IR images, respectively,  was reported in  \cite{[Kang-2014]}.  The \emph{passive} IR band,  the camera sensor detects IR radiation in the form of heat, which is emitted from the face.
 In \cite{[Osia-2017]}, canonical correlation analysis and manifold learning dimensionality reduction is used  in cross-spectral face recognition such as matching visible facial images against images acquired in the passive IR band, and vice-versa.

Detection of the features of plastic surgery \cite{[kn:Bhatt-Plastic-Surgery-2013]} and their mitigation in facial recognition \cite{[Jillela-2012]}, as well as illicit drug and alcohol abuse detection \cite{[Yadav-2016]}, have been investigated for a similar objective. 

These achievements suggest that the technology gap hindering the performance of biometric systems is diminishing.

\subsection{Ambient assistants}\label{subsec:Ambient}

\paragraph{\bf Tasks}
Ambient Intelligent (AmI) systems aim at supporting handicapped and/or elderly people in their daily living activities. The problem addresses  the demographic change in the population and the increase of costs for healthcare.  A survey of AmI technologies is provided in \cite{[Buchmayr-2011]}. 
The core of ambient assisted living is behavioral biometrics \cite{[Tistarelli-2011],[Xefteris-2011]}.

\paragraph{\bf State-of-the-art} 
 Nappi and Wechsler  \cite{[Nappi-Wechsler-2011]} considered biometrics as a means to bridge the gap between \emph{ambient intelligence} (the integration of CI in our natural surroundings)  and \emph{augmented cognition} (the extension of human abilities  in  performance improvement  and  graceful degradation). 
 Design platforms for ambient assisted living  are discussed in \cite{[Phulla-2016]}.
The core technology there has been the wireless body area networks (WBANs) which can monitor
patients or mobile users' health status  \cite{[Viswanathan-2012]}.
 For example, ALADIN (Ambient Lighting Assistance for an Ageing Population) project is based on an adaptive lighting system with intelligent open-loop control to ensure eye health, sleep quality, improved mood, and cognitive performance \cite{[Li-2015]}.

 WBANs provide various services in
different areas such as remote health monitoring, sports, entertainment, and military. It can be used to collect different
physiology parameters including blood pressure, electrocardiography
(ECG), and temperature \cite{[Mitra-2012]}.
Ambient assistants are  crucial components of the smart home \cite{[Cook-2010],[Doctor-2005]} and smart city \cite{[Petrolo-2017]} concept.

The typical ambient intelligence technology gap addresses the passengers of a long flight duration. To reduce possible passenger physical stress and mental distraction,  in-flight exercises have been recommend.  However, the motivation to do these exercises in a confined space during a long-haul flight seems to be a problem. A partial solution, as proposed in \cite{[Westelaken-2011]}, is to equip an airplane seat  with sensors that detect passengers' body movements and gesture. The detected gestures are then used as input for interactive applications in in-flight entertainment systems. A satellite approach is to assess the passenger state using facial expressions.

Another practical application that characterized a significant technology gap addresses the Subjective Well-Being (SWB) problem. This application is based on the CI approaches to measuring the happiness of an individual or group of individuals
 at a particular time, and for a specific activity. SWB is an important indicator of social life and decision-making; it is associated with judgments of how one's life is doing and the amount of positive emotions experienced in one's life.
Instead of the classic question-answer approach, the SWB is based on the automatic monitoring of facial expressions to provide useful support or alternatives. An approach to group happiness assessment using neural networks is introduced in \cite{[Vonikakis-2016]}. The SWB is also a vital component of the monitoring policy in the smart home and smart city concept.

\paragraph{\bf Technology gap}

Adaptation and learning mechanisms are the crucial problems of the ambient assistant development; the feasibility of  solutions to this problem is directly related to the corresponding technology gap. 
A system is said to be adaptive when it is equipped with the ability to respond  to statistical variations of the environment.  Such a system with adjustable parameters  interacts with the environment through sensory inputs and activation outputs. An adaptive element, either internal or external to the system, adjusts the parameters of the system to optimize the performance. This is usually done through a feedback control which refers to an operation that, in the presence of disturbances, tends to reduce the difference between the output of a system and some reference input.

In biometric systems, adaptivity was studied, and some of the technology gaps were identified as shown below.
 Rattani \cite{[Rattani-2015]} defined the task of the 
 adaptation or updating module as a continuous process of adapting the system to the intra-class variation of the input data due to changing acquisition conditions and lifestyle-related changes, \emph{e.g.}, age.
An application of such adaptation can be an e-interviewer that utilizes affective computing that focuses on recognizing, interpreting, and responding to human affects (feelings and emotions) using various sensory modalities such as facial expressions, body language, voice, and other physiological responses \cite{[Calvo-DMello-2010]}. The computer maps measured sensory
data to affective variables such as stress, workload or engagement,
then continuously adapts  its behavior, via a feedback loop, based on the recognized affects. 


In a meta-recognition process for post-recognition score
analysis \cite{[Scheirer-2012],[Scheirer-Meta-Theory-2011]}, the feedback  loop is used for prediction that can take action to improve the overall accuracy of the recognition system. 

An adaptive mechanism is a vital component of any ambient  system. It aims to  regulate the system in order to operate efficiently in dynamic ubiquitous  environments \cite{[Goumopoulos-2011]}.

\subsection{Affect analyzers}\label{subsec:Affect}

\paragraph{\bf Tasks} Biometric-enabled affect analyzers  aim at the detection of patterns  that characterize (directly or indirectly)  the mental and behavior state of a subject  in human-machine and human-human interactions. The core concept of an affect analyzer is a feedback loop in which measured sensory data is mapped into an affective state landscape, such as depression, stress, workload, deception features,  or engagement. In the follow-up interactions, an adaptation  to the detected affects is vital, as adaptation is the fundamental principle of survival of any living organisms. 

A typical example is an affect analyzer for tutoring purposes that aims at the improvement of students' learning.    In this application scenario, automatically detected  contextual cues, facial expressions, and body movements  are fused to make a decision on   confusion, frustration, and boredom. 

\paragraph{\bf State-of-the-art} 
An overview of affect detection methods is given in \cite{[Calvo-DMello-2010],[Zeng-2009]}. 
Affective feedback can be applied to problems such as warning
car drivers in the case of recognized drowsiness \cite{[Picot-2012]}. The classification of a driver's visual attention allocation is of increasing interest in the pursuit of accident reduction. Gaze estimation can be divided into two components: head pose estimation \cite{[Yin-2016]} and eye pose estimation \cite{[Fridman-2016]}.

In \cite{[Monkaresi]},  several sets of features were used in supervised learning for the detection of concurrent and retrospective self-reported engagement: 1) facial expressions from videos, 2) heart rate, and 3) animation units (from the Microsoft Kinect Face Tracker). A Computer Expression Recognition Toolbox (CERT), reported in \cite{[Littlewort-2011]}, can automatically detect Action Units (AUs) as well as head pose and head position information. CERT uses Gabor features as inputs to SVMs that provide likelihood estimates of the presence of 20 different AUs on a frame-by-frame basis. CERT has been tested with databases of both posed and spontaneous facial expressions, achieving accuracies of 90.1\% and 79.9\%, respectively, for discriminating between presence versus absence of AUs.

  The CI is the core of human fatigue detection \cite{[Girard-2014]}. Biometric traits of depression detection (a severe psychiatric disorder preventing a person from functioning normally in both work
and daily lives) is studied in \cite{[Girard-2014],[Wen-2015]}. 
Depression cues can be observed from  audio and  
facial evidence, as well as from body and eye movement. 

The multi-sensory platform called Wize Mirror aims at health related \emph{self-monitoring} and \emph{self-assessment} \cite{[Andreu-2016]}.
 The Wize Mirror detects on a regular basis physiological changes relevant to cardio-metabolic risk factors and offers personalized strategy towards a correct lifestyle, via tailored coaching messages.

A particular type of ambient assistant,  affect cognitive analyzer, and dialog spoken system \cite{[Wu-2015]} is a biometric-enabled interview supporting machine or e-interviewer. The  interview-supporting machines  for security applications aim at deception detection \cite{[Abouelenien-2017],[Nunamaker-AVATAR-2011],[Twyman-2014a]}. It is well documented that in order to achieve an acceptable
practice accuracy for the e-interviewer, there is a need to analyze
and track many modalities containing distinct indicators of
potential deception. Given a question set, responses by an
individual are measured using various biometric modalities
and their relationships. They form the modality-specific information
content or risk deception landscape.

Extended CRFs, the so-called continuous CRFs, have been developed by  Baltrusaitis \emph{et al.} \cite{[Baltrusaitis-2013]} for modeling the affect (emotions from multi-model sources) continuously. 

In the e-interviewer, data acquisition is implemented via
non-contact technologies using various modalities (acquisition
framework) such as the heart rate and blood pressure (in particular, micro color facial changes caused by the heartbeat),
vocalic features, oculometric factors, respiratory functions,
thermal features, and kinesic factors. Analysis and profiling
of the risk of deception requires deep inference technology.
For example, an e-interviewer AVATAR
(Automated Virtual Agent for Truth Assessment in Real-Time)
machine should be smart and introduce itself as a virtual
border officer \cite{[Nunamaker-AVATAR-2011],[Twyman-2014a]}.

Periocular recognition (region around the eye) is of particular interest in watchlist technology and e-interviewers. The authors of  \cite{[Smereka-2015],[Zhao-Kumar-2017]} introduced views on how CI can improve solutions to this challenging problem.

Body odor can be considered as an additional physiological and behavior indicator. For example, some 
moods, such as depression, may affect our body odor.
The odor can also indicate the presence of certain diseases such as skin cancer. 
The most resent results on odor pattern recognition using various CI techniques such as the $k-$nearest neighbors algorithm,  linear discriminant analysis, logistic regression, BNs, and support SVMs are reported in \cite{[Rodriguez-Lujan-2013]}.  

Haarmann \emph{et al.} \cite{[Haarmann-2009]} recognized
the level of arousal from heart rate and skin conductance,
then used it to adapt the turbulence level in a flight simulator. In \cite{[Garbey-2004]},  an approach for the estimation of
blood flow fluctuations from thermal video has been developed. The core is a bioheat transfer model as a partial differential equation  with boundary conditions, that reflects the thermo-physiological processes in a skin region proximal to a major vessel. The inverse problem for this model  is formulated as  the estimation of blood flow speed and related parameters, such as temperature and vessel location of a region of interest (e.g., face), using data from the infrared camera. 

Finally, the problem of critical social importance, that can benefit form using the CI biometric techniques, is support for community and city establishments such as shelters \cite{[OGrady-2011]} and similar facilities that are associated with  social services or  charities  provided to  homeless people, or dislocated due to emergencies or natural disasters.  It is of critical concern that homelessness in today's world is  considered in relation with human dynamics in the changing world, natural disasters, and related potential threats such as epidemics and terrorism.

\paragraph{\bf Technology gap}
Technology gaps in affective system development are related to the following problems \cite{[DMello-2015]}:

\begin{itemize}
		\item Aiming at fusion when collecting and processing data, and in decision-making, shall be a priority, as single biometric modalities may not provide reliable performance. 
			\item  A small subset of the  encompassing modalities and affective states often result in incorrect decisions. 
				\item  Standard procedures are needed for collecting scenarios and the creation of databases (benchmarks) for the evaluation and comparison of affect detectors.
\end{itemize}

Crowdsourcing is a general name for techniques that involve posing
many small-scale tasks to a crowd of  users, and piecing together
the crowd's responses to achieve a larger-scale goal. 
Computer games are a primary example of where computational actions can be adjusted to the player’s facial expression \cite{[Mcduff-2012]}. However, interest in facial expressions has moved far beyond computer vision and interaction research, reaching areas like media consumption or health applications. 

Another example is driver distraction and inattention leading to crashes and incidents. In \cite{[Rigas-2012]}, the driver's stress is studied using a BN.
For this, the electrocardiogram, electrodermal activity, and respiration, as well as past observations of driving behavior are used. This study can be useful as a baseline (ground truth) for the approaches based on distant measures. Results on driver's eye/head tracking reported in \cite{[Ahlstrom-2012]} can be useful in comparison with the same scenario but processing via CI tools. 

In health  self-monitoring and self-assessment (Wize Mirror concept) \cite{[Andreu-2016]}, the CI related gaps are identified using 1) the scientific challenge (intelligent methods for translating face signs into repeatable,
accurate, and efficient computational measures), and 2) the technological challenge (such as a non-intrusive platform, seamlessly integrated into a daily-life environment, temporal and spatial data synchronization, and real-time
processing of multimodal data).

\subsection{Synthesizers of biometric traits}\label{subsec:Synthesizers}

\paragraph{\bf Tasks}  Generating artificial  or synthetic biometric traits (\emph{e.g.}, face, fingerprints, iris, signature) that are as real as possible. Various approaches are known. Consider, for example, the  reconstruction paradigm. Let the model be chosen,  its input and output is represented by a biometric trait  and its extracted features, respectively. Now the  task is  to check whether a model extracts a reconstruction of the input itself from the features.  Let  the PCA model be chosen. In this model, the mapping of the feature space
is a linear projection into the leading principal components
and the reconstruction is performed by another linear projection. It is predictable, that the reconstruction is perfect only for those data points that lie in the linear subspace spanned by the leading
principal components which are  the structures captured by this model.

There is a way for improvement of the above approach. In PCA for instance, the input is reconstructed  from the features to assess the quality of the encoding, while in a probabilistic setting we can analyze and compare different models in terms of their conditional probability distributions.  These reconstructions can be more like real data compared, for example, with  PCA.

\paragraph{\bf State-of-the-art}
Two challenging problems are recognized in  deep learning techniques: 
1) the amount of 
 data required to populate/train the  parameters, and
2) the duration of training.
The solution of the first and second problem addresses the development of synthesizers of biometric traits and a parallel algorithm with hardware acceleration support (e.g. \cite{[Yuan-2017]}), respectively.

Cappelli \cite{[Cappelli-2003]} was the first to develop a commercially available fingerprint synthesizer (software package) based on an augmentation paradigm  for the purpose of evaluating fingerprint recognition algorithms.  Any biometric trait can be augmented using techniques such as translation, rescaling, cropping, and others, to design a synthetic trait.  
Any CI method can be used for this purpose, such as genetic algorithms  \cite{[Cho-2007]};  one has to start using a set of real fingerprint acquisitions, then apply an evolutionary or other algorithm to initialize a set of filters, which are used to modify the fingerprint images and generate synthetic samples. 

The control of generated traits  is a key problem in biometric synthesizer design. In particular, the initial data can contain features of interest and the synthesizer should generate a set of traits with these features. In this content,  the method proposed in \cite{[Zhao-2012]} provides some additional possibilities via control of the statistical models of the features.

A review of biometric synthesizers and their applications  according to the following  taxonomical view can be found in \cite{[Yanushkevich-2007]}:
i) synthesis as an inverse problem of biometrics, and
ii) synthesis in terms of biometric data forgery.
 The underlying paradigm is defined as \emph{analysis-by-synthesis}, or \emph{learning-by-synthesis}, that is, synthesis-based feedback control.  

Note that inverse problems are typically ill-posed tasks. In addition to synthetic biometric traits generation, an inpainting of the face images \cite{[Sulam-2016]} was formulated as an inverse problem with a sparse-promoting prior based on the learned global model; a dictionary learning technique was used that utilized adaptive learning on a set of atoms representing real signals as sparsely as possible.

Banerjee \emph{et al.} \cite{[Banerjee-2017]} have formulated the 
requirements for a facial synthesizer for the need of training deep inference engines, such as CNNs:
i) generate an arbitrary large number of facial images,
ii) and generate a balanced number of images per person.
This also allows for  avoiding any  potential issues of invasion of privacy.
The paper  reported on successful experimental results on training the CNN for face recognition using 200,000 synthetic faces with a resolution of $512\times 512$.

In summary, the  progress achieved to date in synthetic biometrics  is as follows:  
\begin{itemize}
\item []\hspace{-8mm}  \emph{Synthetic 2D fingerprints:} Initial data in the Capelli \cite{[Cappelli-2003]} synthesizer include fingerprint type, image size, region of interest, and singular points.  In the approach developed in \cite{[Zhao-2012]}, statistical features of a master fingerprint are used to generate multiple impressions. 
 The possibilities of a synthesizer based on an evolutionary paradigm have been studied in \cite{[Cho-2007]}.
\item []\hspace{-8mm}  \emph{Synthetic 3D fingerprints:} 
Kuecken \cite{[Kuecken-2004]} developed the biologically inspired mathematical models of  fingerprint formation.  
In \cite{[Labati-2013]}, a virtual environment  was developed for  generating 3D fingerprint samples;  the synthesizer  starts  from a real or synthetic image of contact-based acquisition.
\item []\hspace{-8mm}  \emph{Synthetic hand prints:} 
The hand print is a hybrid high-resolution biometric trait that combines fingerprints, the palmprint, and hand-shape. A synthesizer of hand prints is proposed in  \cite{[Morales-2014]}. This is an example of a synthetic multi-biometric.
\item []\hspace{-8mm}  \emph{Synthetic hand geometry:}  In  \cite{[Gomez-Barrero-2012]}, this task is formulated as 
a reverse engineering (inverse biometric) problem: recover the
original hand geometry sample from its parametrized template. Such a reconstruction process has been generally referred to as inverse biometrics.
\item []\hspace{-8mm}  \emph{Synthetic iris:}  The PCA, MRF, and   Gabor filter are well suited for iris synthesis \cite{[Zuo-2007]}.
The main concern of the recent report \cite{[Cardoso-2013]} is the simulation of uncontrolled acquisition conditions.
\item []\hspace{-8mm}  \emph{Synthetic 3D eye:} In \cite{[Wood-2016]}, an eyeball and eye region synthesizer have been developed for training a gaze estimator based on a CNN.
\item []\hspace{-8mm}  \emph{Synthetic face:} It is a common practice to augment any facial database, using techniques such as  translation and rotation, as well as CNNs \cite{[Banerjee-2017]}.  In security, synthetic facial traits, such as morphing images,  are identified as attacks. Therefore, new databases and detectors of such attacks shall be developed.
\item []\hspace{-8mm}  \emph{Facial sketches:} This is a particular case of synthetic facial images.   Survey  \cite{[Wang-N-2014]}, as well as papers \cite{[Zhang-S-2017],[Zhang-L-2015]}, provide a necessary platform for facial sketch techniques.  
A synthesizer that explores the differences in
 visible face images and  thermal images was proposed in \cite{[Osia-2017]}.  
A study by Berger \emph{et al.} \cite{[Berger-2013]} contributes to a more complicated problem, -- both style and abstraction in sketching of a human face. Klum \emph{et al.} \cite{[Klum-2014]}  studied how facial sketches can be used in law enforcement,  in particular, in watchlist checks at mass-transit hubs.
\item []\hspace{-8mm}  \emph{Synthetic signature:} A 3D model of synthetic signatures for biometric system training was first suggested in \cite{[Yan-2005]}. In \cite{[Galbally-2012]}, an approach based on a priori knowledge about a certain signature have been  developed. A set of parameters of required signatures are input data to this synthesizer.   It is different from  the augmentation-based techniques that starts from one or more real samples of a given person.

\item []\hspace{-8mm}  \emph{Synthetic gait:} Gait synthesizers using HMM and PCA models are studied in \cite{[Tilmanne-2012]}.  An augmentation paradigm was chosen in \cite{[Han-2004]} for synthetic gait template design using real gait energy images.  
\item []\hspace{-8mm}  \emph{Synthetic speech:} It is a  common practice to use the CI techniques in $i)$ speech synthesizers  for the purpose of training speaker verification systems \cite{[Tokuda-2013]}, as well as $ii)$ in speech spoofing detectors  \cite{[Sanchez-2015]}.
\item []\hspace{-8mm}  \emph{Synthetic body:} Body images captured beyond the visible spectrum (the range of $30-300$ GHz that corresponds to wavelengths $10-1$ mm) overcome  some limitations such as clothes variations. A database of millimeter wave body images (shapes) was created in  \cite{[Moreno-2010]}. Millimeter band biometric traits, such as respiration, heartbeats, voice, gait, body shape \cite{[Petkie-2009]},  are usually studied in combination with concealed weapon detection.

\end{itemize}

The above is a brief introduction to synthetic biometrics. Our goal  is to highlight the urgent unsolved problems in this area. 

\paragraph{\bf Technology gap}

The following CI-related technology gaps and emerging problems are identified in the area of synthetic biometrics.
\begin{itemize}
\item [--]  \emph{Biases:} Synthetic biometric data are characterized by various biases that are difficult or impossible to predict.  In particular, Galbally \emph{et al.} \cite{[Galbally-2014]} proposed a set of measures  for the detection of differences in image quality properties between real and synthetic biometric traits. 
\item [--] \emph{Deep inference:} The deep inference engine is needed for exploring high order dependencies in quality of  real and synthetic biometric traits. For example, differences can address the optical processes of 3D biometric traits acquisition versus the technique of 2D synthetic traits.
\end{itemize}
It should be noted that the results obtained using existing databases of synthetic biometric traits have to be used with caution.  Synthetic traits may not reflect the performance of a recognition system in a real-life scenario.   For example, it was reported in \cite{[Gottschlich-2014]}, that synthetic features can be detected in a popular Capelli's fingerprint synthesizer  \cite{[Cappelli-2003]}. A second order minutiae statistic reflecting the covariance structure of the minutiae distributions was applied for that in \cite{[Gottschlich-2014]}.
In terms of countermeasures, synthetic biometric data can be used to train the biometric system to recognize spoofing.

\section{Summary and conclusions}\label{sec:Summary-conclusion}

This survey is motivated by ever raising requirements for the performance of biometric-enable systems in many emerging  applications. We identify and analyze the existing challenges using technology gap methodology \cite{[Hartman-Technology-Gap-Analysis-2005]}. An adequate response to these challenges inevitably involves CI techniques such as machine learning, in particular, deep inference, and  advanced pattern recognition paradigms.  The  reason is that  biometric traits of interest are often are hidden in high order dependencies of patterns,  while the power of many popular statistical techniques is limited and prevents the proper dealing with such data.

To the best of our knowledge, this survey is the first attempt, after Frey and  Jojic \cite{[Frey-2005]}, to revise  available  CI-related resources for deep inference needs. Conceptually, our study is a practical reflection of the mentioned study with an extension via Hinton's team's results on deep inference \cite{[Hinton-Sejnowski-1986],[Hinton-2006],[Hinton-2006-a],[Salakhutdinov-2012]}.  Recently, Wang and Yeung \cite{[H-Wang-2016]}  introduced their vision on the development of Bayesian deep learning models.   A synergy of these works is reflected in this survey on CI-related techniques and cognitive technology in biometrics.

While the efficiency of the most  CI-related techniques and methodologies has been proven in various fields time and again, the  inference for biometric applications requires a much more detailed study and mathematical modification.   For example, message-passing inference  provides near information theoretical limits in communication, such as turbo encoding,   but this border  performance is not straightforward  in biometric-enable applications. The latter  strongly depends on the task,  scenario, and available data \cite{[Frey-2005],[Zheng-2012]}.

 The key conclusions from our study are as follows:

1)  Emerging  applications demand ever increasing 
 performance in  terms of recognition accuracy, reliability of decision-making, fusion at all levels, adaptivity, countermeasures, and robustness. It also calls for   bridging  with breakthrough achievements in forensic and cognitive studies.  The CI-related techniques provide promising results in this domain. 

2) \textbf{New inference and learning algorithms are required for system level decision-making.}  The high priority techniques include \emph{e.g.}, DBNs, message-passing inference (Bayesian networks and MRFs).   The concept of the inference engine that combines different inference paradigms, is a promising prospect because it explores the uncertainty in an aggressive way (different interpretations of uncertainty and multi-metrics). Because of the high complexity of such inference, hardware acceleration, such as VLSI architecture for the RBM \cite{[Yuan-2017]}, can be useful in real-time applications.
 
3) \textbf{Techniques for unsupervised training of deep inference engines is the high priority problem.} It is a commonly accepted fact that the possibilities of biometric benchmark databases which contain natural biometric traits, are limited for the purposes of deep inference learning, because it is difficult or impossible to achieve an acceptable training quality. The development of  the generators of synthetic biometric traits is a reasonable alternative. 


Finally, the  mass-transit systems and border agencies worldwide are looking towards integrating biometrics into  Blockchain Emergency (BE) identification, which is important in times of humanitarian crises and natural disasters.  Pilot Projects on BE e-ID are currently being conducted in several countries, in particular, by Canadian \cite{[Canada-Blochain-2016]}  and US \cite{[DHS-Blochain-2015]} Border Agencies.

\subsection*{Acknowledgment}
This project was partially supported by  Natural Sciences and Engineering Research Council of Canada (NSERC) through the grant ``Biometric intelligent interfaces''.

\end{document}